\documentclass{article}

\usepackage{microtype}
\usepackage{graphicx}
\usepackage{subcaption}

\usepackage{booktabs} 

\usepackage{hyperref}

\usepackage[accepted]{icml2023}

\usepackage{amsmath}
\usepackage{amssymb}
\usepackage{mathtools}
\usepackage{amsthm}

\usepackage[capitalize,noabbrev]{cleveref}

\theoremstyle{plain}

\theoremstyle{definition}

\theoremstyle{remark}

\usepackage[textsize=tiny]{todonotes}

\icmltitlerunning{Can We Understand Plasticity Through Neural Collapse?}

\begin{document}

\twocolumn[
\icmltitle{Can We Understand Plasticity Through Neural Collapse?}



\icmlsetsymbol{equal}{*}

\begin{icmlauthorlist}
\icmlauthor{Guglielmo Bonifazi}{equal,eth}
\icmlauthor{Iason Chalas}{equal,eth}
\icmlauthor{Gian Hess}{equal,eth}
\icmlauthor{Jakub Łucki}{equal,eth}
\end{icmlauthorlist}

\icmlaffiliation{eth}{Department of Computer Science, ETH Zurich, Zurich, Switzerland}

\icmlcorrespondingauthor{Jakub Łucki}{jlucki@student.ethz.ch}

\icmlkeywords{Machine Learning, Deep Learning, Plasticity Loss, Neural Collapse}

\vskip 0.3in
]



\printAffiliationsAndNotice{\icmlEqualContribution} 

\begin{abstract}
This paper explores the connection between two recently identified phenomena in deep learning — plasticity loss and neural collapse. We analyze their correlation in different scenarios, revealing a significant association during the initial training phase on the first task. Additionally, we introduce a regularization approach to mitigate neural collapse, demonstrating its effectiveness in alleviating plasticity loss in this specific setting.
\end{abstract}

\section{Introduction}
\label{introduction}

Neural plasticity is a term originating from neuroscience, which refers to the ability of a nervous system to adapt itself in response to new experiences \cite{bioneuralplasticity}. Recently, this notion has been extended to artificial neural networks (NN), 
since it has been observed that they suffer from plasticity loss (PL) \cite{understandingplasticity}.
Several works have noted that NNs trained on a non-stationary objective underperform on new tasks, compared to NNs trained only on the new task starting from randomly initialized weights \cite{primacy_bias, lyle2022understanding}. 
\\
However, retraining a NN for each new task is sometimes infeasible. For instance, in continual learning an agent is required to learn indefinitely \cite{dohare2023loss}. Similarly, in reinforcement learning settings the relation between inputs and prediction targets changes over time \cite{lyle2022understanding}. Lastly, recommender systems often experience popularity distribution shifts, which harm their performance \cite{popularity_shift}. \\
As a consequence, multiple works have attempted to understand PL better and prevent it. It has been conjectured that several NN's properties such as weight norm, feature rank or number of dead units might serve as an explanation for the loss of plasticity \cite{conjecture_plasticity}. However, \citealp{understandingplasticity} argue that none of those properties fully explain loss of plasticity, and identify the smoothness of the loss landscape as a possible factor. 

Another recently observed phenomenon, termed \textit{neural collapse} (NC), has sparked interest in the deep learning community \cite{neuralcollapse, han2021neuralcollapse, zhu2021geometric}. We hypothesize that NC might compromise the plasticity of a neural network. Our hypothesis is based on the following considerations. Firstly, NC involves multiple interconnected phenomena, one of them being the collapse of last-layer features to their respective class means \cite{neuralcollapse}.
This becomes evident in the terminal phase of the modern training paradigm, when further reducing the loss even after achieving a zero classification error \cite{neuralcollapse, han2021neuralcollapse}. 
In the context of achieving perfect classification on a fixed task, this is advantageous, as it enhances linear separability of classes. \\
However, we hypothesize that an excessive reduction in in-class variability of activations might compromise the plasticity of the model. Indeed, the collapse of features to class means implies a form of \textit{forgetting}, where information distinguishing samples within a class is discarded. \citealp{achille2019critical} describe such a \textit{forgetting} phase after which the ability of the network to change its effective connectivity decreases.\\
Secondly, the periodic reinitialization of the last few layers, where NC occurs, has been shown to prevent overfitting to early experiences in reinforcement learning \cite{primacy_bias}. Finally, both PL and NC seem to exacerbate with the number of training steps taken \cite{lyle2022understanding, neuralcollapse}.\\
To the best of our knowledge the relationship between neural collapse and plasticity loss has not been investigated in the scientific literature yet. 

\section{Models and Methods}
\label{methods}

We investigate the relation between PL and NC in two different continual learning settings in which PL has been observed before: Permuted MNIST \cite{dohare2023loss} and Warm starting \cite{ash2020warmstarting}. Although initially planned, we did not conduct experiments in the Continual Imagenet setting \cite{dohare2023loss} due to resource constraints. The varying difficulty levels of tasks learned in sequence necessitate a larger number of runs for meaningful conclusions, requiring substantial computational resources.

In all experiments we measure the four neural collapse metrics introduced in \citealp{neuralcollapse}, but focus mostly on NC1, which measures cross-example within-class variability of last-layer features. Specifically, we measure $Tr(\Sigma_W\Sigma_B^\dag/C)$, where $Tr(\cdot)$ is the trace operator, $\Sigma_W$ and $\Sigma_B$ are the within-class and between-class covariances of the last-layer activations of the training data, respectively, $C$ is the number of classes and $[\cdot]^\dag$ is the Moore-Penrose Pseudoinverse. The lower the NC1 metric, the more collapsed are the last-layer features.

\subsection{Permuted MNIST}
This experiment based on classical MNIST entails a series of tasks. Each task is characterised by pixel permutation that is applied to all images within the dataset. Following the principles outlined in \cite{dohare2023loss}, we conducted experiments utilizing a 3-layer Multi-Layer Perceptron (MLP). This choice was motivated by the loss of spatial information due to pixel permutation, rendering Convolutional Neural Networks (CNNs) ineffective in this particular scenario. Because of computational constraints, we chose an MLP configuration with $100$ units in each hidden layer, a setup considered sufficient for the simplicity of the task. Moreover, we opted for a step size of $0.01$, as this selection showcased the greatest plasticity loss according to \cite{dohare2023loss}. In our pursuit to investigate the relationship between plasticity loss and neural collapse, we deliberately altered the training duration for the first task in experiments 2 and 3. For all subsequent task we use only one epoch.

We tested the following experimental setups:
\begin{enumerate}
    \item \textbf{Neural collapse in continual learning:}\\
    We sequentially train on 140 distinct tasks, assessing neural collapse and plasticity loss following the completion of each task.
    \item \textbf{Variable length of training on the first task:}\\
    We experiment with varying the number of epochs only in the initial task. Subsequently, we assess the correlation between neural collapse in the first task and the plasticity loss in the following tasks.
    \item \textbf{Variable neural collapse in the first task:}\\
    This experiment is analogous to the previous setting, however here we are training a model on the first task until its NC1 decreases below a threshold. We used thresholds of $0.22, 0.20, 0.18$ to ensure different NC1 values and repeated the experiment for 5 seeds. Furthermore, we measure the number of epochs required to go below the threshold. This allowed us to check the correlation between neural collapse, plasticity loss and time of training.
    
\end{enumerate}

\subsection{Warm starting}
We employed the CIFAR-10 dataset, utilizing two distinct versions: one comprising half of the samples (referred to as \textit{Warm-Up} dataset) and another containing the entire dataset (referred to as the \textit{Full} dataset). In accordance with \cite{ash2020warmstarting}, we initiated training with an 18-layer ResNet on the \textit{Warm-Up} dataset for a variable number of epochs. Subsequently, starting from the checkpoint of this warm-up training phase, the model underwent further training on the \textit{Full} dataset until a training accuracy of 99\% was achieved—this convergence criterion was used in all subsequent experiments.
\\
We conducted the following experiments:
\begin{enumerate}

    \item \textbf{Correlation in classical warm up}:\\
   We assessed neural collapse on the \textit{Warm-Up} dataset using four distinct metrics. Subsequently, we explored the correlation between neural collapse and test accuracy on the full dataset.
   \vspace{-4 pt}
\item \textbf{Correlation under weights shrinkage and perturbation}:\\
   We shrink and perturb the weights on the warmed-up model.  More technically, taking $\Theta$ as the set of parameters, $n$ as the number of units and $\varepsilon \sim \mathcal{N}(0, \frac{2}{n})$,  we have: $\forall \theta \in \Theta, \hspace{5 pt}\theta_{\text{sp}} = \lambda \theta + B \varepsilon$. This technique was used in \cite{ash2020warmstarting} to enable faster training and better generalization on the subsequent task. Following this modification, we measured neural collapse on the adjusted weights and utilized them as a new checkpoint for training on the \textit{Full} dataset. The correlation between neural collapse and test accuracy on the \textit{Full} dataset, as influenced by the modified weights, was then investigated.\vspace{-4 pt}
\item \textbf{Warm up training with Neural Collapse Regularization:}\\
    To assess if neural collapse has a causal effect on plasticity loss we intervened on it by adding a NC1 regularization term on the loss during the warm up phase. We then measured the effects of this regularization on the test accuracy of both the \textit{Warm-Up} dataset and the full dataset.
\end{enumerate}

All experiments were run on 5 different seeds to ensure reproducibility and to present the uncertainty of our claims. Details on hyper-parameters can be found in the appendix, while the code used to obtain the results is available at the following GitHub repository: \url{https://github.com/gianhess/dl_project.git}. 

\section{Results}
\label{results}

\subsection{Permuted MNIST}
\begin{figure}[!h]
  \centering
  \begin{subfigure}{\linewidth}
  \centering
    \includegraphics[width=0.75\linewidth]{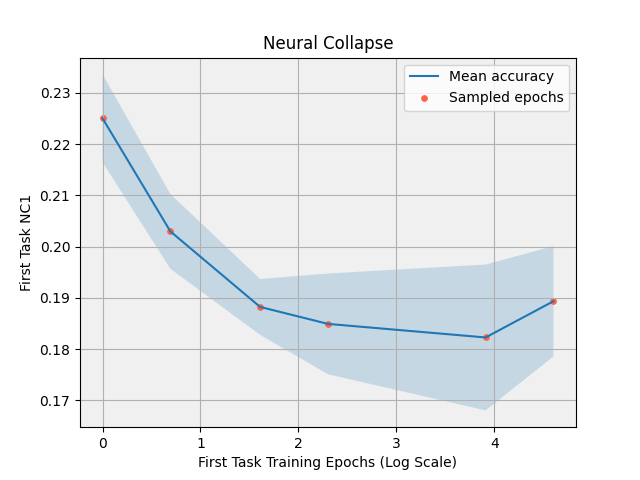}
    \label{fig:img1}
  \end{subfigure}
  \vspace{-0.2cm}
  \begin{subfigure}{\linewidth}  
  \centering
    \includegraphics[width=0.75\linewidth]{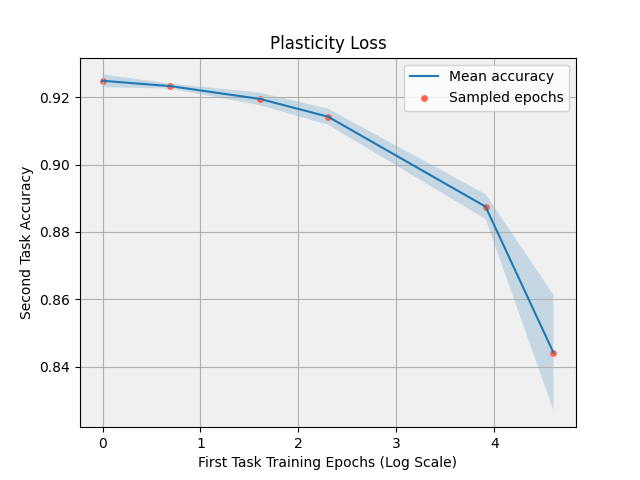}
    \label{fig:img2}
  \end{subfigure}
  \caption{NC1 and plasticity loss for different initial task training epochs [1,2,5,10,50,100]. Shaded regions are obtained adding and subtracting the standard deviation to the mean.}
  \label{fig:pm_nc_pl}
\end{figure}

\begin{enumerate}
    \item \textbf{Neural collapse in continual learning:}\\
    Figure \ref{fig:nc_acc} clearly shows the loss of plasticity as the training accuracy decreases with the task index. Furthermore, we observe that NC1 metric is increasing with the task index. The two metrics are highly correlated as indicated by Pearson correlation coefficient of $-0.94$. Therefore, we can conclude that the plasticity loss prevents neural collapse from occurring. This follows from the definition of NC1 since with higher plasticity loss the number of misclassifications is larger and thus the within-class variability increases.
    \begin{figure}[!h]
      \centering
      \includegraphics[width=0.4\textwidth]{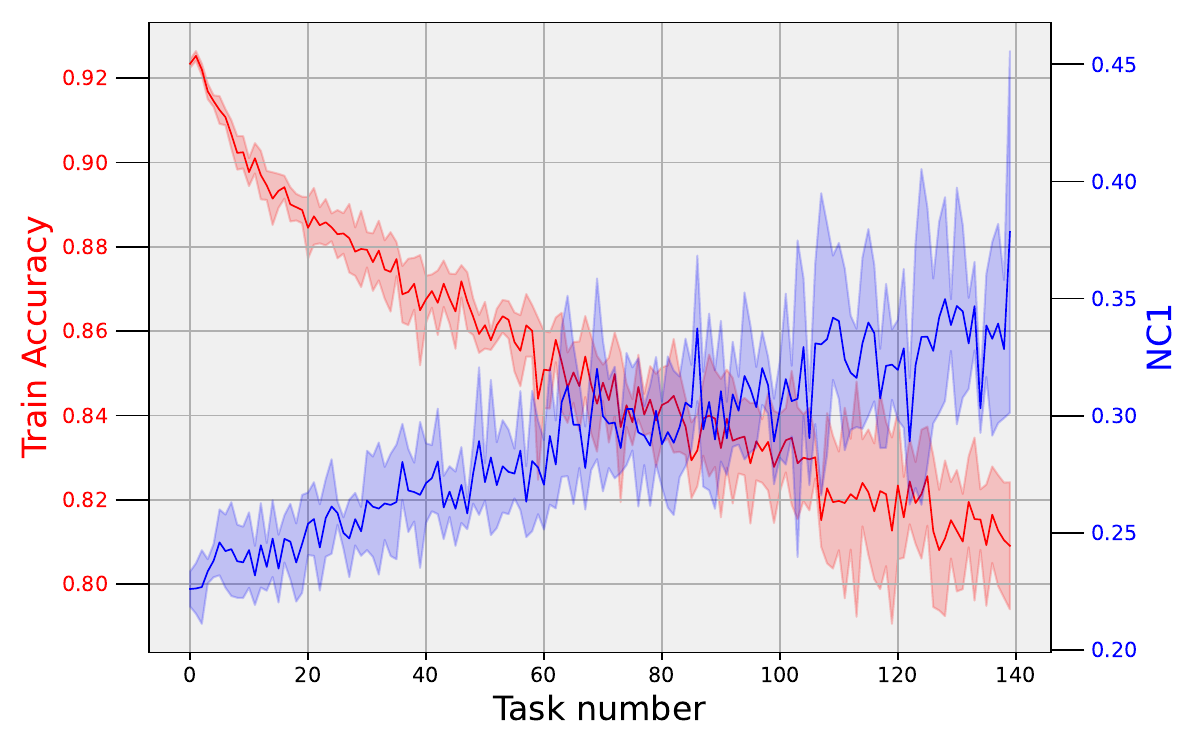}
      \caption{NC1 and Training Accuracy in the continual learning setting for 140 tasks. Error bars indicate the standard deviation calculated based on 6 different seeds.}
      \label{fig:nc_acc}
    \end{figure}
    \item \textbf{Variable length of training on the first task:}\\
        
    In Figure \ref{fig:pm_nc_pl}, it is evident that, the value of NC1 initially decreases as the first task is trained on more epochs. However, beyond a certain threshold, it reaches a plateau and no further neural collapse is observable. On the other hand, there is a significant correlation between the loss of plasticity and the number of epochs used for training the initial task. Consequently, an observable correlation between NC1 and plasticity loss emerges initially. However, this correlation is not perfect, as beyond a certain threshold, we witness a constant neural collapse and a simultaneous escalation in plasticity loss.
    
    \item \textbf{Variable neural collapse in the first task:}\\
    We have obtained 14 different NC1 values after the first task. The Pearson correlation coefficient of this metric with the number of epochs required to reach the threshold is $-0.99$. On the other hand the correlation coefficient between NC1 and the performance on the second task is $0.60$. Hence, it is possible that the underlying cause for both neural collapse and plasticity loss is based on the length of training and the there is no causal dependency between the two.
\end{enumerate}
\vspace{-0.6cm}

\subsection{Warm starting}
The insights derived from Experiment One and Experiment Two lead to the following key observations:
\begin{itemize}
\item At the beginning, there exists an exceptionally strong positive correlation between neural collapse and plasticity loss (test accuracy on the test dataset). In simpler terms, a higher degree of neural collapse corresponds to a significant increase in plasticity loss.
\item This correlation experiences rapid deterioration as training progresses. To illustrate this phenomenon, we computed correlations within a sliding window of 100 warm-up epochs.
\end{itemize}

\begin{figure}[!h]
  \centering
  \includegraphics[width=0.4\textwidth]{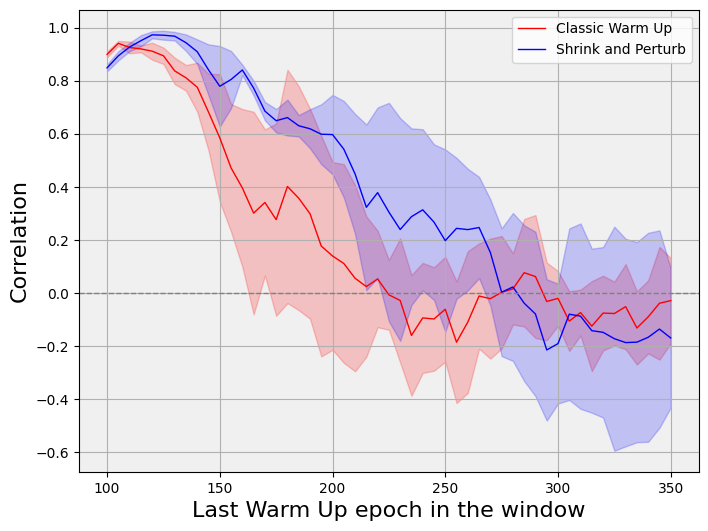}
  \caption{Correlation between NC1 and test accuracy on \textit{Full} dataset with a window size of 100. Shaded regions are obtained adding and subtracting the standard deviation to the mean.}
  \label{fig:corrWU}
\end{figure}

Figure \ref{fig:corrWU} shows this behavior for NC1, clearly demonstrating it with both normal and shrink-and-perturbed weights.
showing the relation of all neural collapse metrics to PL can be found in Appendix \ref{nc_evolution}.

Given the robust correlation observed in the early stages of warm-up training, we investigated the impact of neural collapse regularization. The results indicated a marked improvement over classical warm-up, reflected in better test accuracy for both the \textit{Warm-Up} and \textit{Full} datasets. This suggests that, at least within this experimental setting, plasticity loss can be alleviated by regularizing neural collapse, mitigating the decline in test accuracy on a new task resulting from prior training on a different task.

\begin{figure}[!h] 
  \centering
  \includegraphics[width=0.48\textwidth]{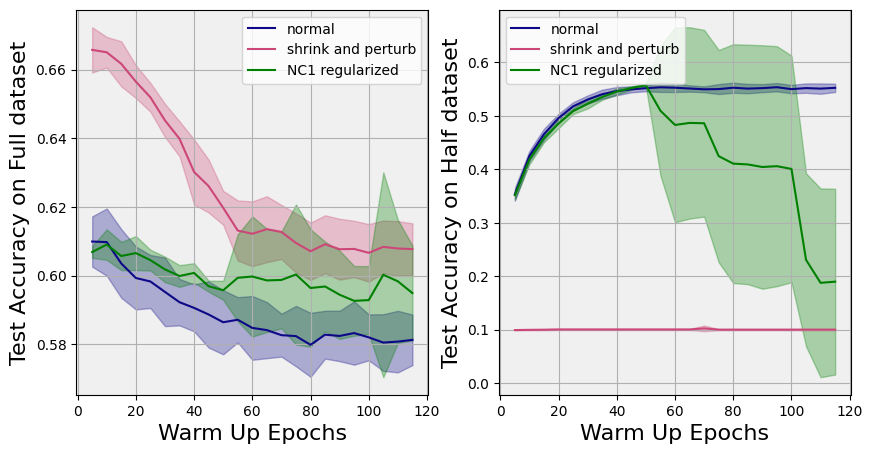}
  \caption{Test accuracies on \textit{Warm-Up} dataset (Right) and \textit{Full} dataset (Left) for different warm up schemes. Shaded regions are obtained adding and subtracting the standard deviation to the mean.}
  \label{fig:accWU}
\end{figure}

In Figure \ref{fig:accWU} we compare this result to the test accuracies achieved after shrinking and perturbing the weights of the warmed-up model. While higher accuracies are achieved, it is important to note that in contrast to NC1 regularization, applying shrink-and-perturb requires knowledge of when a task change will occur. Furthermore, with NC1 regularization, performance on the \textit{Warm-Up} dataset is maintained if training is stopped after convergence, while the performance of the shrinked-and-perturbed model on the \textit{Warm-Up} dataset completely deteriorates.

\section{Discussion}
\label{discussion}
The interplay between Plasticity Loss (PL) and Neural Collapse (NC) is intricate, influenced by mutual dependencies and external factors. Our investigation with the Permuted MNIST, has yielded two key observations.

Firstly, in a continual learning scenario, the onset of PL implies an inability to fit subsequent tasks, essentially preventing the model from reaching neural collapse by definition. This is underscored by the negative correlation depicted in Figure \ref{fig:nc_acc}.

Secondly, a nuanced interpretation arises when focusing on the post-first-task-change phase, where is possible to achieve neural collapse. Here, varying the number of epochs or the attained NC reveals a pronounced positive correlation, suggesting that higher levels of NC correspond to increased PL on the second task. However, caution is warranted in drawing definitive conclusions, as at least at the beginning training duration is inherently correlated with both NC and PL, possibly acting as an underlying factor driving their observed correlation.

Consistent with the latter finding are the outcomes from the Warm Starting experiment. There, a strong and positive correlation between the two is evident, even though it diminishes as training on the initial task progresses. Notably, in this context, we were even able to successfully leveraged NC to influence PL, hinting at the potential for a causal relationship between the two phenomena. We should also mention that other regularization methods like L2 could still have the same beneficial effect on PL.

As highlighted earlier, numerous variables influence the relationship between PL and NC. Factors such as network size and optimization schedules determine the model's capacity to overfit on the first task. Additionally, the similarity between subsequent tasks can lead to divergent outcomes, such as the one witnessed with plasticity loss or a successful fine-tuning. Given this complexity, any claims about the relationship should be made cautiously. This project, however, has provided valuable insights into the intricate aspects underlying PL and NC, emphasizing the need for thorough exploration of these variables in future studies.

\section{Summary}
\label{summary}
This paper aimed to analyse the relation between neural collapse and plasticity loss. We have considered two different settings and we carried out different experiments to check for correlation and potential causal link between the two. Our findings are that in continual settings as soon as we experience plasticity loss, neural collapse cannot be present. Considering cases in which we are able to overfit on the first task, we can instead observe a strong correlation between the two, which nonetheless diminish as training on the first task progresses. Therefore, our primary contribution lies in experimental verification of a relation between the two phenomena.

\nocite{langley00}

\bibliography{report}
\bibliographystyle{icml2023}

\newpage
\appendix
\onecolumn
\section{Warm Starting}

\subsection{Evolution of Neural collapse metrics and Plasticity Loss over the number of warm-up epochs}

\label{nc_evolution}

\begin{figure}[!h]
  \centering
  \begin{minipage}{0.3\textwidth}
    \centering
    \includegraphics[width=\linewidth]{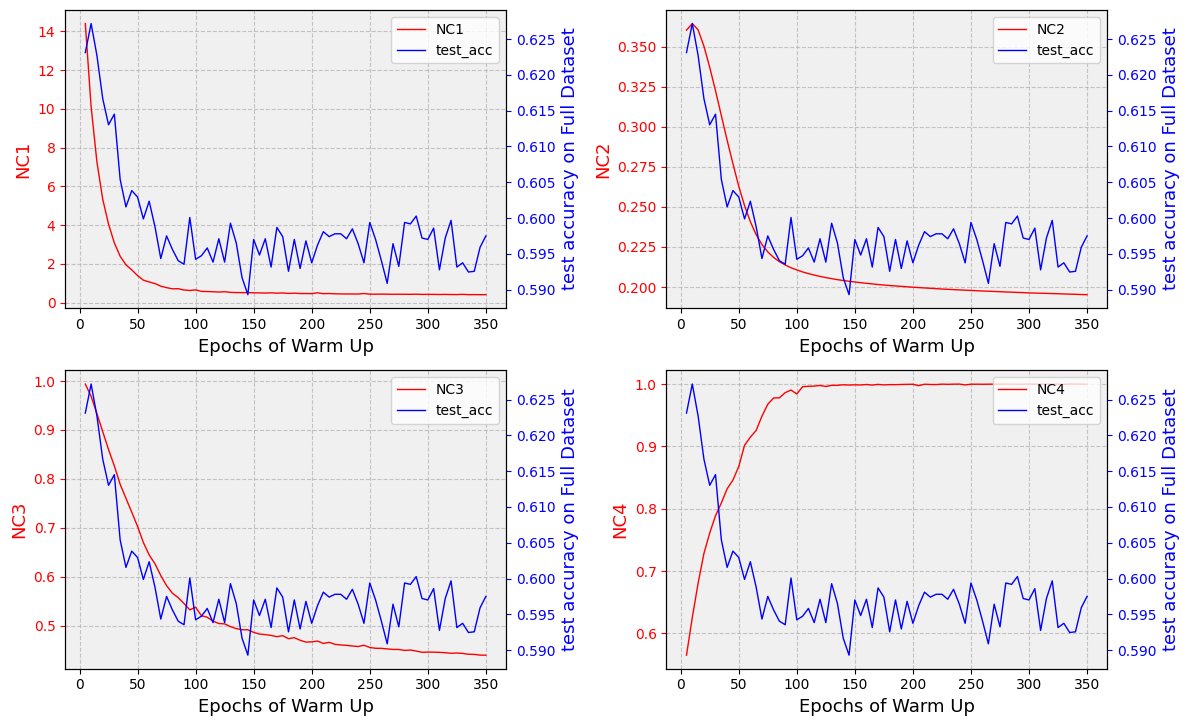}
    \caption{NC metrics against test accuracy on \textit{Full} dataset across all warm-up epochs. }
    \label{fig:img1}
  \end{minipage}%
  \hfill
  \begin{minipage}{0.3\textwidth}
    \centering
    \includegraphics[width=\linewidth]{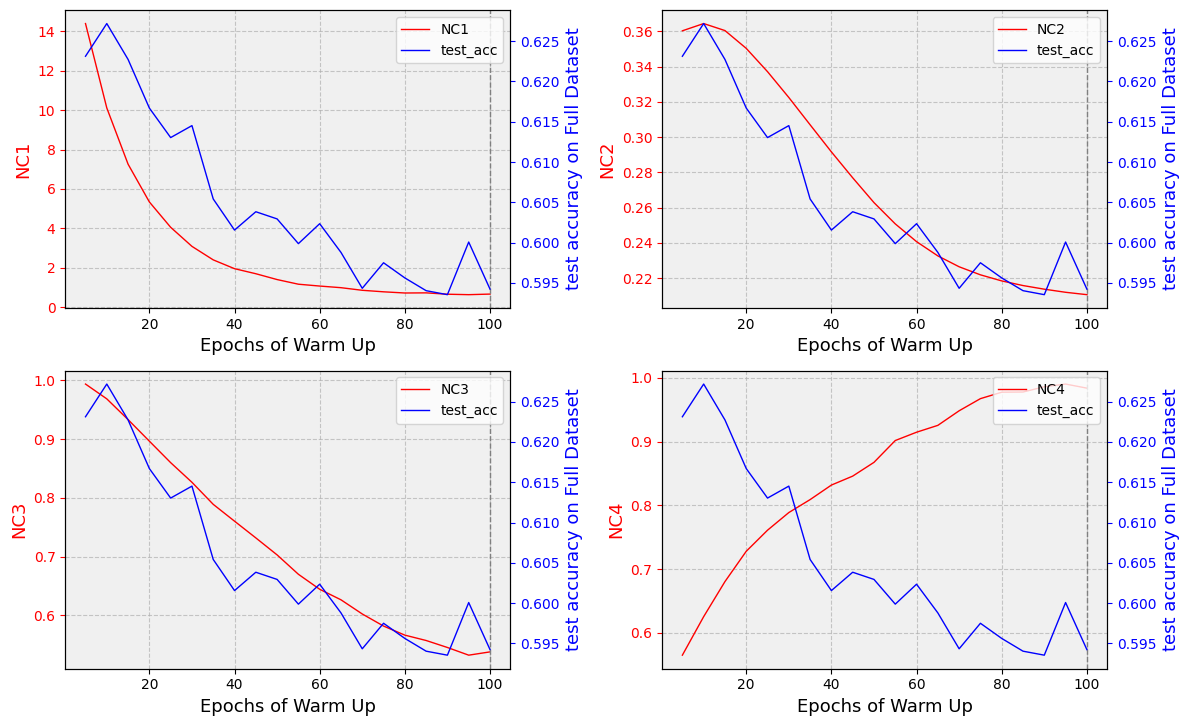}
    \caption{NC metrics against test accuracy on \textit{Full} dataset across the first 100 warm-up epochs. We can observe a clear correlation with all of them.}
    \label{fig:img2}
  \end{minipage}%
  \hfill
  \begin{minipage}{0.3\textwidth}
    \centering
    \includegraphics[width=\linewidth]{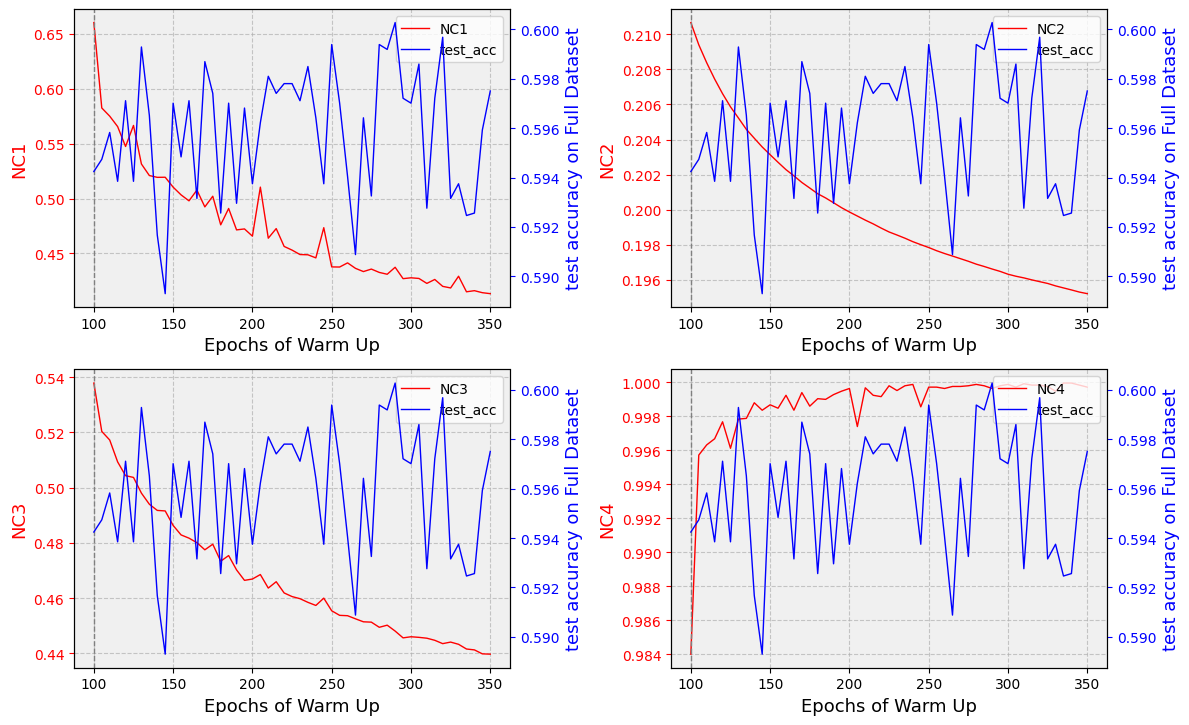}
    \caption{NC metrics against test accuracy on \textit{Full} dataset after the first 100 warm-up epochs. No correlation is present at this stage, NC continues to exacerbate while test accuracy oscillates around a constant value.}
    \label{fig:img3}
  \end{minipage}
  \label{fig:overall}
\end{figure}

\subsection{Hyper-parameters for Warm Starting.}
Following \cite{ash2020warmstarting} in all the experiments we used a 18-layered ResNet as model and SGD with 0.001 learning rate and no momentum. For Shrink and perturb we used $\lambda = 0.6$  and $B = 0.01$. Finally we used a weight factor of 0.05 in the NC1-regularization, defining the loss as $\mathcal{L} = \mathcal{CE} + 0.05 \cdot NC1$, with $NC1$ computed batch-wise and $\mathcal{CE} $ being the cross-entropy loss.


\end{document}